\def\year{2022}\relax
\documentclass[letterpaper]{article}
\usepackage{aaai22} 
\usepackage{times} 
\usepackage{helvet} 
\usepackage{courier} 
\usepackage[hyphens]{url} 
\usepackage{graphicx} 
\usepackage{graphicx} 
\usepackage{natbib} 
\usepackage{caption} 
\DeclareCaptionStyle{ruled}%
{labelfont=normalfont,labelsep=colon,strut=off}
\usepackage{array,ragged2e}
\newcolumntype{P}[1]{>{\RaggedRight\arraybackslash}p{#1}}
\usepackage{graphicx}
\usepackage{wrapfig}
\usepackage{epsfig}
\usepackage{xcolor}
\usepackage{float}

\usepackage{hyperref}

\begin{document}
%
\title{Robin: A Novel Online Suicidal Text Corpus of Substantial Breadth and Scale}
\author{Daniel M. DiPietro\equalcontrib\footnote{dipietrodaniel131@gmail.com}, Vivek Hazari\equalcontrib, Soroush Vosoughi\textsuperscript{\rm \ddag}
}

\affiliations {
    \textsuperscript{\rm \ddag} Dartmouth College, Department of Computer Science
}

\maketitle


\begin{abstract}
\begin{quote}
Suicide is a  major public health crisis. With more than 20,000,000 suicide attempts each year, the early detection of suicidal intent has the potential to save hundreds of thousands of lives. Traditional mental health screening methods are time-consuming, costly, and often inaccessible to disadvantaged populations; online detection of suicidal intent using machine learning offers a viable alternative. Here we present Robin, the largest non-keyword generated suicidal corpus to date, consisting of over 1.1 million online forum postings. In addition to its unprecedented size, Robin is specially constructed to include various categories of suicidal text, such as suicide bereavement and flippant references, better enabling models trained on Robin to learn the subtle nuances of text expressing suicidal ideation. Experimental results achieve state-of-the-art performance for the classification of suicidal text, both with traditional methods like logistic regression (F1=0.85), as well as with large-scale pre-trained language models like BERT (F1=0.92). Finally, we release the Robin dataset publicly\footnote[1]{Available for download \href{https://drive.google.com/file/d/1MRGis3s4RQ2MMSQu2pxqBqOzStRCeN86/view?usp=sharing}{$<$here$>$}.} as a machine learning resource with the potential to drive the next generation of suicidal sentiment research.
\end{quote}
\end{abstract}


\section{Introduction}
\noindent In 2017 alone, approximately 800,000 people died from suicide globally \citep{nimh_stats}. Suicide has become an epidemic--within the United States, it now ranks as the second leading cause of death for those aged 10 to 34 and the tenth leading cause of death overall \citep{nimh_stats}. Since 75\% of those that commit suicide were not known to have a mental health condition in the four weeks preceding their death, early identification of risk factors has the potential to save hundreds of thousands of lives each year; suicide prevention interventions offer a statistically significant improvement in patient outcomes \citep{ahmedani2014health, liu2019proactive}. Unfortunately, current screening methods are time consuming and costly, requiring trained medical professionals. To complicate matters further, there is a severe mental healthcare shortage in the United States, with experts estimating that the current number of mental healthcare professionals must nearly quadruple to meet demand \citep{kff_2020}.

The rise of social media may offer a solution. Through social media, billions of users disseminate self-writing on a daily basis. The primary users of social media--young people--are those at greatest risk for suicide. However, applying natural language processing for online screening is no easy task; signs of suicidal ideation in self-writing can be subtle and are often incredibly nuanced. Flippant references to suicide are common and semantically similar to genuine ideation, complicating matters even further.

In order to build robust machine learning models that can accurately detect suicidal sentiment online, large, comprehensive datasets of suicidal text are essential. Previous works have used datasets that are lacking in size or generated using simple techniques such as keywords \citep{shing2018expert, roy2020machine, sinha2019suicidal, liu2019proactive, ji2018supervised}. At this time, to the best of our knowledge, there is no publicly available corpus of suicidal text that offers the breadth and scale needed to drive further natural language processing research in this problem.

In this paper, we present a novel suicidal text corpus called Robin. Robin consists of over 1.1 million scraped social media posts, roughly 220,000 of which express suicidal intent; for comparison, this number is nearly two orders of magnitude higher than previous datasets that were not generated using keywords. Additionally, Robin was specifically curated to reflect the various categories of suicidal language, such as flippant references to suicide and suicide bereavement; human validation demonstrated high inter-annotator agreement and highly accurate labeling within the dataset. As a proof of concept, both traditional models, such as logistic regression, and large scale pre-trained language models, such as BERT, achieved state-of-the-art performance for the classification of suicidal text when trained using the Robin dataset. These results demonstrate the potential to drive the next generation of and natural language processing research in the detection of suicidal sentiment online, a problem with massive implications for society.


\section{Related Work}

\paragraph{Research on Suicidal Sentiment Online}

Past psychological research has presented methods of understanding and classifying suicidal sentiment. \citet{ellis1988classification} describes four categories or 'dimensions' of suicidcal phenomena: descriptive, situational, psychological/behavioral, and teleological. 

Researchers \citet{burnap2015machine} present a series of expert-developed categories which represent common forms of communication on the topic of suicide. These include 'Evidence of Possible Suicidal Intent', 'Campaigning', 'Flippant Reference', 'Information or Support', 'Memorial or Condolence', and 'Reporting of Suicide'. The breadth of these categories has proved to provide a challenge for understanding online suicidal sentiment.
    
\paragraph{Machine Learning for Mental Healthcare and PSPO} 

Psychiatry has long been explored as a powerful avenue for the application of machine learning techniques. \citet{srividya2018behavioral} and \citet{durstewitz2019deep} provide surveys of some of the applications of machine learning and deep neural networks (DNN)s in psychiatry, which include the diagnosis and prognosis of neuroimaging data, and predictions on patient health data in large datasets. \citet{durstewitz2019deep}  also points out that one of the challenges in applied deep learning (DL) to psychiatry is that datasets are often very small, and that in this domain data is usually very high dimensional relative to the typical sample sizes collected in neuroscientific studies.

Suicide notes, often studied via applied machine learning techniques, have been a major research subject as well \citep{huang2007hunting, spasic2012naive, pestian2010suicide, yang2012hybrid, liakata2012three, desmet2013emotion, ghosh2021multitask}. However, the most actionable data would come from those who have expressed suicidal ideation but have not yet committed suicide.

Among many researchers investigating the application of machine learning to suicidal ideation, \citet{liu2019proactive} outlines a new paradigm of applied psychiatry called ``Proactive Suicide Prevention Online'' (PSPO). PSPO outlines a paradigm where machine learning models identify online users who express suicidal sentiment, and support is provided to them directly. Their results were promising but limited to Chinese internet users.

\paragraph{Existing Suicidal Text Datasets}

Many pre-existing suicidal corpora use heuristics to approximate clinical truth, such as the content that users choose to post on social media. Twitter is a common source of suicidal text, with researchers \citet{burnap2015machine, ji2018supervised, roy2020machine, sarsam2021lexicon, rajesh2020suicidal, sinha2019suicidal} creating Twitter corpora ranging in size from a few hundred to a few million tweets. Due to the nature of the Twitter API, all of these datasets are generated by querying for keywords that the researchers have determined are related to suicide topics. As a result, models trained on these datasets perform extremely well in classification tasks, as the tweets which are suicidal contain pre-selected keywords, which are easily discernible by machine learning models. Further, the terms of the Twitter API means that it is impossible for researchers to share their datasets, so every researcher attempting to investigate suicidal sentiment data must regenerate their own dataset \citep{twitterdeveloper}.

Other researchers have have tried compiling survey data to determine risk of suicide \citep{srividya2018behavioral, nobles2018identification, majumder2021machine}. This approach often coincides with high quality annotation and validation, as these surveys are conducted by clinical psychologists who are experienced with suicidal risk. Still, it is challenging for these types of datasets to reach a scale where machine learning techniques can perform effectively. 

A third approach is to compile posts from online forums dedicated to suicidal topics. Researchers in \citep{ji2018supervised, tadesse2020detection, liu2019proactive, cheng2017assessing, cao2019latent, shing2018expert, falcone2020digital} have taken this approach. \citet{ji2018supervised} compiled a notable dataset of 7201 total posts, with 3549 suicidal posts originating from a forum dedicated to expressing suicidal sentiment, /r/SuicideWatch. This approach uses the source of the data to label whether or not the data is suicidal. This is advantageous when compared to approaches which use keywords relating to suicide ideation, as it does not introduce any biases or pre-conceptions that the authors compiling the dataset may have towards what constitutes suicidal sentiment. In a work using data collection methods similar to those presented in this paper, \citet{shing2018expert} compiled a dataset of 11,129 Reddit users who had posted in /r/SuicideWatch at least once and randomly sampled 934 from them to form a time-series dataset of posts; support vector machine models trained on this dataset achieved an F1-score of 0.66 in the binary classification task.


\section{The Robin Dataset}

\subsection{Collection and Preprocessing}

The Robin dataset is a suicidal text corpus designed to offer state-of-the-art breadth and scale, with the intention of enabling next-generation suicide prevention machine learning models. With this in mind, the dataset consists of 1,104,711 online posts, collected from a variety of subreddits from the social media website Reddit.com. PushShift is an open source API which allows users of the API to powerfully search and aggregate submissions and comments from Reddit. PushShift was used to greatly reduce the collection time of the Robin dataset \citet{baumgartner2020pushshift}.

Constructing a dataset using data from Reddit brings numerous benefits. First, outside of a few online acronyms, posts on Reddit have relatively few structures in place that differ from those commonly employed in the English language at large. As a result, the Robin dataset can be used for the machine learning detection of suicidal content, rather than suicidal content limited to a single platform. Datasets constructed using other platforms, such as Twitter often contain artificially shortened self-writing and website-specific acronyms and abbreviations; consequently, they may struggle to generalize to the typical English employed on websites like Facebook and Reddit. All Reddit posts are publicly available, which makes scraping them relatively easy compared to websites where profiles may be private, such as Facebook or Instagram.

Reddit is divided into countless subreddits, which are individual online communities where users post about specific topics. We curated data from several subreddits, with an underlying rationale for each. Suicidal posts make up 20\% of the dataset and were sourced from the subreddit SuicideWatch, which allows users to post a cry for help when they are feeling suicidal thoughts. We operate under the assumption of self-labeling, assuming that all posts on SuicideWatch are expressing suicidal ideation or sentiment; human annotation later on demonstrates that this is a quite reasonable assumption. We included 13 additional subreddits, each specifically selected to either offer an additional category of suicidal text or to provide general, non-suicidal text. Suicidal text is nuanced and spans a range of categories, only one of which--suicidal ideation--is cause for concern. Incorporating these additional categories, like suicide bereavement, allows models using Robin to train on common false positives. The categories of suicidal text included in Robin and their subreddit sources are described in detail in Table \ref{table:rationale}.

\begin{table*}
\centering
\begin{tabular}{{p{2cm}p{3.5cm}p{9cm}}}
\hline
\textbf{Suicidal Category} & \textbf{Subreddit Source(s)} & \textbf{Explanation}\\
\hline
Ideation & SuicideWatch & Suicidal ideation describes the feeling or desire of wanting to commit suicide; all ideation posts were sourced from SuicideWatch. Users post on SuicideWatch as a cry for help when they are at risk of suicide; thus, these posts indicate suicidal ideation.\\
Flippant & TIFU, CasualConversation, self, teenagers & Flippant references to suicide employ suicidal language for a rhetorical or emphatic effect but do not express genuine suicidal intent.\\
Bereavement & SuicideBereavement & Suicide bereavement expresses grief related to another person's death from suicide, but these posts themselves do not express genuine suicidal intent.\\
Support & CasualConversation, self, teenagers & These posts express support for those affected by or experiencing suicidal thoughts, but these posts do not express suicidal intent themselves.\\
Awareness & MensRights, CasualConversation & These posts are intended to raise awareness for the issue of suicide, but they do not express genuine suicidal intent themselves.\\
News & SuicideBereavement, CasualConversation & These posts record public suicides, such as celebrity suicides, but they do not express genuine suicidal intent themselves.\\
\hline
\end{tabular}
\caption{\label{table:rationale}
Robin includes six categories of suicidal text, only one of which expresses genuine suicidal ideation. Note that each subreddit may contain posts spanning several categories of suicidal text. Additionally, if a subreddit is listed, it does not imply that all posts from that subreddit (with the exception of SuicideWatch) necessarily fall into any given suicidal category.}
\end{table*}

Robin was constructed using the web API service PushShift, which allows users to query specific subreddits and download available posts from as early as the beginning of 2019. For each post, we collected the title, body, subreddit, length, and identifying id. The specific breakout of the dataset by subreddit is shown in Table \ref{table:breakout}.

Relatively little pre-processing was needed. First, the title and body were concatenated to create a single unit of text; this was done to make the corpus platform-agnostic, as many platforms do not have post titles. Then, each post was assigned a suicidal/non-suicidal label, and the reddit post IDs were replaced with dataset-specific identifiers. Finally, we removed subreddit-specific features not reflecting the syntax of plain English. This was necessary for the subreddit tifu, in which the body of posts always begins with the token TIFU.

\begin{table*}[h]
\centering
\begin{tabular}{llllll}
\hline
\textbf{Subreddit} & \textbf{Label} & \textbf{Posts} & \textbf{Percentage} & \textbf{Authors} & \textbf{Mean Length}\\
\hline
    books & 0 & 7,155 & 0.65 & 6,094 & 788\\
    casualconversation & 0 & 220,384 & 19.95 & 83,982 & 503\\
    jokes & 0 & 7,155 & 0.65 & 4,832 & 156\\
    mensrights & 0 & 37,140 & 3.36 & 15,285 & 1,202\\
    politicaldiscussion & 0 & 40,675 & 3.68 & 14,906 & 795\\
    relationships & 0 & 110,471 & 10.00 & 78,422 & 2,385\\
    self & 0 & 151,420 & 13.71 & 65,109 & 1,006\\
    showerthoughts & 0 & 71,555 & 0.65 & 5,364 & 12\\
    suicidebereavement & 0 & 3,305 & 0.30 & 2,044 & 1,283\\
    suicidewatch & 1 & 220,941 & 20.00 & 116,995 & 949\\
    teenagers & 0 & 110,470 & 10.00 & 30,651 & 206\\
    television & 0 & 7,155 & 0.65 & 4,890 & 556\\
    tifu & 0 & 166,972 & 15.11 & 110,493 & 1,587\\
    truegaming & 0 & 14,313 & 1.30 & 7,487 & 1,526\\
    \textit{total} & N/A & 1,104,711 & 100.00 & N/A & 1,030\\
\hline
\end{tabular}
\caption{\label{table:breakout}
Robin is made up of data from 14 subreddits, totalling approximately 1.1 million posts. Note that the ``Percentage'' column refers to how much of the Robin dataset each subreddit accounts for. Additionally, the mean length column refers to the mean number of characters per post.}
\end{table*}

\subsection{Dataset De-Identification}

Whenever handling mental health data, it is of the utmost importance that adequate steps have been taken to ensure anonymity and privacy. Although Reddit usernames are rarely self-identifying, we made the decision not to include them as a feature in the dataset given the sensitive nature of this work. However, while the posts were being scraped, we tracked the number of unique authors on each subreddit. Additionally, each Reddit-specific post identifier was replaced with a dataset-specific post identifier, creating an additional barrier separating the scraped posts in Robin from their online Reddit versions. It’s worth noting that these posts are publicly available online. Still, we took the aforementioned steps to ensure that they are anonymized in the context of the Robin dataset. All scraping and post collection was done in compliance with the Reddit Terms of Service and Conditions.

\subsection{Annotation and Validation}

Data labeling was conducted under the assumption that all posts on SuicideWatch are suicidal and all posts from the other subreddits are non-suicidal. In order to assess our quality of labeling, we randomly selected 1000 posts from the dataset for human annotation with an even split of 500 suicidal and 500 non-suicidal. Each post was annotated by five annotators using Amazon's MTurk service, with a fee of \$0.03 per post (corresponding to a wage of around \$18 per hour). Annotators were given the following prompt: `Is the author of this social media post suicidal?'. Since each post is labeled by five annotators, we select the label with the majority of votes, with no chance of a tie. Each post was labeled by five annotators, but these may not be the same five annotators for each post. 

Of the 500 posts labeled as suicidal in the sample, a majority of annotators classified 398 as suicidal and 102 as non-suicidal. Of the 500 posts labeled as non-suicidal in the sample, a majority of annotators classified 487 as non-suicidal and 13 as suicidal. These results yield an annotator F1-score of 0.87 and accuracy of 0.89, indicating fairly close agreement between our labeling and the annotator labeling. Additionally, there was considerable inter-annotator agreement, with an average Fleiss' Kappa of 0.78. If anything, these labeling results indicate that Robin is overly biased in labeling content as suicidal. That said, false positives on this problem are far less impactful than false negatives.

While, on the one hand, this can be used as validation for our existing labeling, on the other hand, it can also be used to assess the performance of untrained humans on this task, assuming that our original labeling was accurate. This is probably not a perfectly true assumption, but it can offer insight nonetheless. Some of the suicidal posts that were labeled as non-suicidal by the annotators can well be seen as suicidal given the correct context.

\subsection{Linguistic Analysis}

We conducted an extensive linguistic analysis of the Robin dataset, examining the most common bigrams and trigrams in each subreddit, as well as the frequency of suicidal terms.

\begin{figure}[h]
    \centering
    \includegraphics[width=0.45\textwidth]{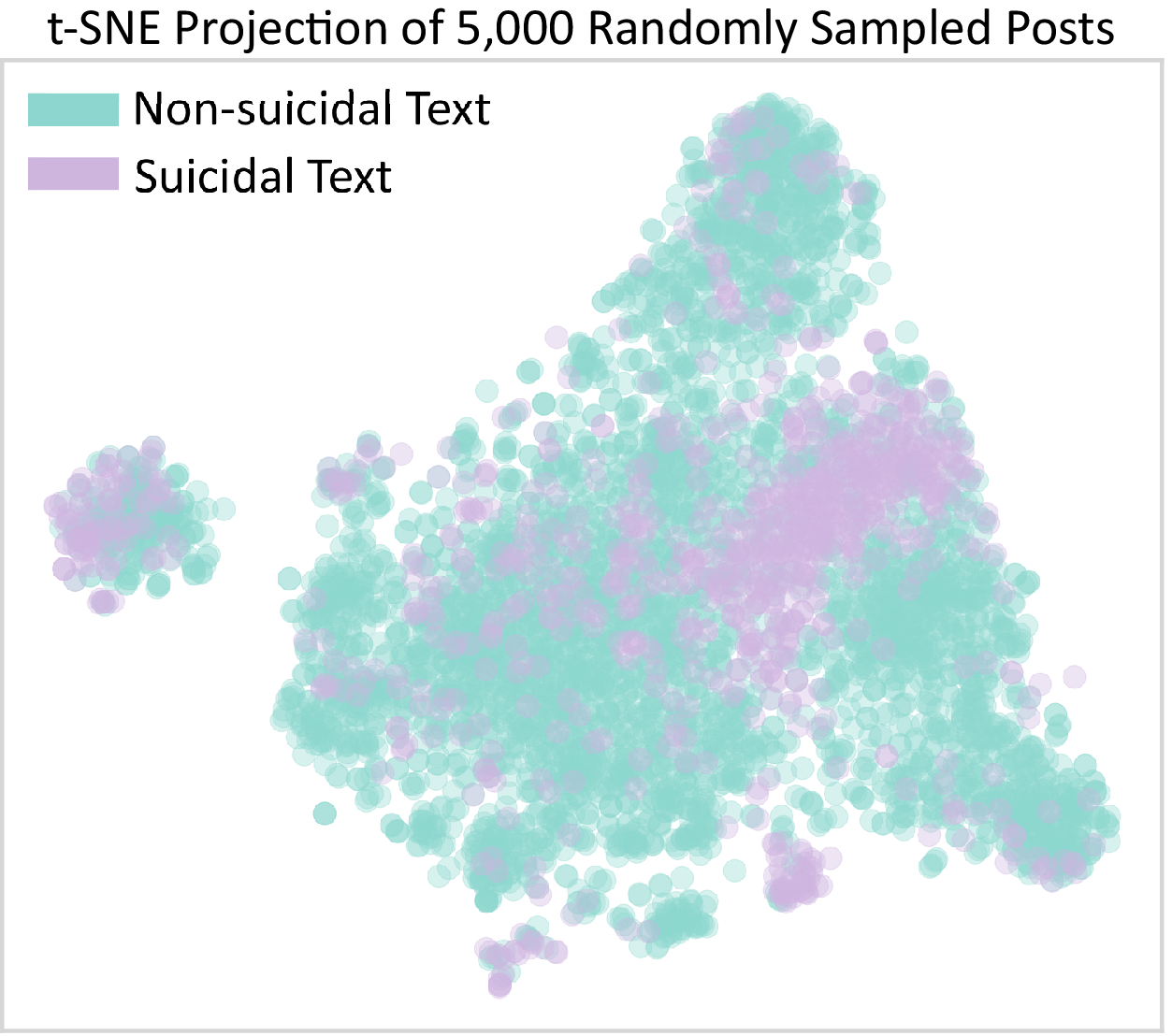}
    \caption{This t-SNE projection indicates that there exists some distance between suicidal and non-suicidal text in its original dimensionality; however, the scattering of suicidal text demonstrates that the classification task is non-trivial when keyword-bias is removed.}
    \label{fig:tsne}
\end{figure}

We also conducted a t-SNE dimensionality reduction to visualize a random sample of 5,000 posts from our dataset to better understand the structure of our data. We used TF-IDF vectorization for this subset of the data. Figure \ref{fig:tsne} shows that the lexical content of the suicidal data tends to have some distance from other types of posts, but is scattered throughout the visualization. This suggests that a purely lexical model may not be adequate to distinguish suicidal from non-suicidal text.

\begin{table*}[h]
\centering
\begin{tabular}{P{2cm}P{1cm}P{1.5cm}P{4cm}P{5cm}}
\hline
\textbf{Paper Presented} & \textbf{Source} & \textbf{Size} & \textbf{Validation} & \textbf{Notes}\\
\hline
    
    \citet{shing2018expert} & Reddit & 1,868 users & Partial expert annotation (26.2\%), crowd-sourced annotation & Every post of users who had posted in SuicideWatch was collected and labelled suicidal  \\
    
    \citet{roy2020machine} & Twitter & 7,223,922 tweets & Tweets matching specific phrases & No human annotation, generated by querying for keywords  \\
    
    \citet{sinha2019suicidal} & Twitter & 34,306 tweets & Tweets matching keyword search & Keywords generated by analyzing SuicideWatch  \\
    
    \citet{liu2019proactive} & Sina Weibo & 12,786 posts & Annotation by psychology post-grads & Posts collected from a wall of an influential micro-blogger who commited suicide  \\
    
    \citet{ji2018supervised} & Reddit, Twitter & 7,201 posts, 10,288 tweets & Researcher annotated with rule `expressing suicidal thoughts' & Reddit posts collected from SuicideWatch, tweets collected with keywords  \\
    
    This Paper & Reddit & 1,104,711 posts & Observed user behavior, crowdsourced annotation & Dataset is composed based on subreddits from which posts originate  \\
\hline
\end{tabular}
\caption{\label{table:comparison}
A comparison of notable suicidal datasets.}
\end{table*}

\subsection{Comparisons to Selected Prior Work}

We present a brief overview of notable previously published datasets in Table \ref{table:comparison}. 

\citet{shing2018expert} presented a corpus which contained a time-series of posts for 1868 Reddit users. They constructed their corpus by finding 11,129 users which had posted at least once in SuicideWatch and had at least 10 Reddit posts between January 1, 2008 and August 31, 2015. Then, they randomly sampled 1,097 users from this. Of the 1,097, they were able to successfully annotate the posts of 934 users. To form the overall dataset, they included an equal number of control users who were not suicidal. Although this dataset uses a similar data collection methodology to ours, the size of the corpus they present is over an order of magnitude smaller than ours. Further, they excluded all users with less than 10 posts. Users on SuicideWatch often create a new account which they intend to use once to post on SuicideWatch. Omitting this type of post leaves the corpus with an incomplete picture of suicide ideation.

\citet{roy2020machine}, \citet{sinha2019suicidal} both generated datasets via the Twitter API. The Twitter API only allows users to collect tweets that match specified search queries. As a result, these datasets only captured tweets which contain semantically obvious suicidal intent, i.e. ``I am thinking about killing myself.'' This approach is flawed for two reasons. First, there is an inclusion bias against tweets which express suicidal ideation but do not contain obvious keywords like `suicide' and `die'. Second, selecting Tweets by keywords alone over-simplifies the problem of suicidal sentiment classification (allowing bag-of-words models to excel by virtue of how the dataset is constructed) and injects a major bias for what constitutes suicidal text. 

We want to emphasize that there is a dearth of comparable corpora that are publicly available. We emailed authors of related work for dataset access, and no emails were returned. We believe this further highlights the need for the Robin corpus: note that the Robin dataset is freely and publicly available under the Creative Commons BY 4.0 license. It is available for download \href{https://drive.google.com/file/d/1MRGis3s4RQ2MMSQu2pxqBqOzStRCeN86/view?usp=sharing}{$<$here$>$}.

The authors are aware of and acknowledge several limitations in our dataset compared to other work. Although our validation achieved high inter-annotator agreement, our annotators were online workers based in the United States, not experts. The size of the dataset meant that the 1,000 posts which were randomly sampled to be validated comprise a small fraction of the total dataset. Finally, our data originates from anonymous Reddit postings. This introduces bias in the demographics for which our model is finely tuned towards the male, English-speaking demographic group with which Reddit is associated \citep{tigunova2020reddust}.


\section{Benchmarking}

\begin{figure*}[hbt!]
    \centering
    \includegraphics[width=0.95\textwidth]{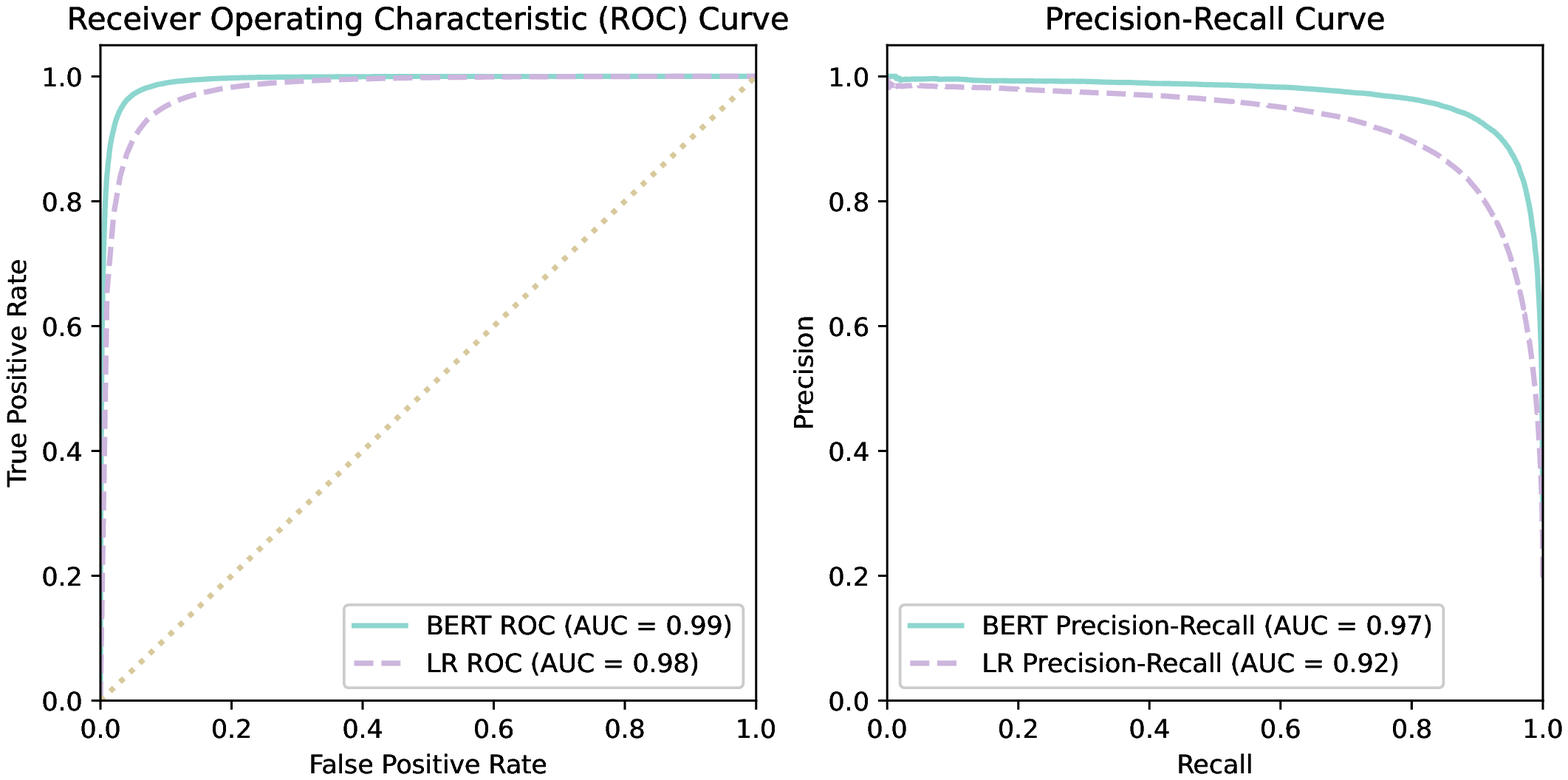}
    \caption{Receiver Operating Characteristic (ROC) and Precision-Recall Curves for logistic regression and fine-tuned BERT models trained on the Robin dataset.}
    \label{fig:roc_lr}
\end{figure*}

\subsection{Overview and Methods}

Although researchers are increasingly emphasizing deep learning techniques, traditional machine learning techniques such as logistic regression still offer impressive performance for many tasks. To this end, we examined a breadth of traditional learning methods to establish a comparative benchmark for subsequent deep learning approaches. We trained the following models: logistic regression, linear support vector machines, complement and multinomial Naive Bayes, random forest, and XGBoost. Each model was hyperparameter tuned using F1-score and cross-validated on five folds of data. Hyperparameter tuning grids were relatively coarse, especially for tree-based approaches such as random forest and XGBoost, due to environmental and computational concerns (see Appendix for exact tuning details).

Each traditional learning model was paired with six different vectorization approaches, with hyperparameter tuning happening for all six pairs. The six vectorization approaches used were unigrams, bigrams, and unigrams and bigrams; each was used with either bag-of-words vectorization or a term frequency-inverse document frequency (TF-IDF) transformation. For traditional machine learning approaches, we performed vectorization within each fold. Additionally, all standard English stop words were removed, and the vocabulary was capped at 20,000 unique words.

Once all traditional models were adequately tuned, the best model was selected for further analysis based on F1-score. In this additional analysis, we benchmarked the model using several additional metrics, measured subreddit-specific performance, generated receiver operating characteristic (ROC) and precision-recall curves, and examined linguistic markers by measuring the feature importance embedded within the model.

We also  applied the BERT transformer model \citep{devlin2018bert}. BERT pre-trained models can be fine-tuned to specific tasks, allowing for noteworthy performance on particular objectives while remaining context-aware due to BERT's extensive pre-training on BooksCorpus and English Wikipedia. Our results use a fine-tuned pre-trained BERT model ('bert-base-uncased'), implemented in the library Transformers by HuggingFaces, with a 768-node fully connected classification and softmax layer appended onto the end \citep{wolf2020transformers}. We truncated posts to 512 tokens, and padded any posts which were shorter than this length to meet the formatting requirements of BERT. We used a tokenizer from Transformers to pre-process the text to a BERT-specific format, and used the popular Adam Optimizer for our training \citep{kingma2014adam}. We trained BERT on several subsets of our dataset and continued to see an increase in performance as the size of the dataset we used continued to grow. When initially fine-tuning BERT, we tried a learning rate of 0.01, which quickly led to over-fitting. We finalized with a learning rate of $5 \cdot 10^{-6}$ and batch size of 10 over a single training epoch. Due to computational constraints when benchmarking BERT's performance, we did not use cross validation and used no tuning besides learning rate, whereas we used five-fold nested cross validation for traditional machine learning techniques. For traditional machine learning approaches, we performed vectorization within each fold. We split our dataset into an 80-20 train-test split for traditional machine learning and for BERT. The random seed we used for model training was 42. 

\subsection{Traditional Machine Learning Classification}

After hyperparameter tuning, the best-performing model was logistic regression paired with TF-IDF vectorization of unigrams and bigrams; this architecture yielded an F1-score of 0.85 ($\pm 0.002$) when tested on approximately 200,000 unseen posts. Other techniques also offered noteworthy performance, with even the worst-performing model achieving an F1-score of 0.73 ($\pm 0.001$).

Beyond F1-score, the logistic regression model achieved a precision of 0.89, recall of 0.82, accuracy of 0.94, and Matthews' Correlation Coefficient (MCC) of 0.82. The model cumulatively classified 861,092 true negatives, 180,191 true positives, 22,670 false positives, and 40,747 false negatives across all of the cross-validation folds. The ROC and Precision-Recall curves demonstrate the model's excellent capability in distinguishing between classes, achieving AUCs of 0.98 ($\pm 0.000$) and 0.92 ($\pm 0.000$) respectively (Figure \ref{fig:roc_lr}).

We also examined the linguistic markers of the logistic regression model by charting the fifteen most strongly weighted features as determined by coefficient magnitudes in Appendix Table \ref{fig:importance_lr}. While some terms, such as `suicidal' and `kill' are obvious linguistic markers, others, such as `anymore' and `goodbye' are far more subtle. Many of these terms may not indicate suicidal ideation depending upon the context, emphasizing potential shortcomings of such naive language models.

\subsection{Large Scale Language Model Classification}

As expected, BERT was able to learn the nuances of suicidal text and outperformed the traditional methods in every measured metric. When evaluated on roughly 200,000 test posts, the model classified 36,939 suicidal posts as suicidal, 3,288 suicidal posts as non-suicidal, 156,440 non-suicidal posts as non-suicidal, and 3,332 non-suicidal posts as suicidal, achieving an F1-score of 0.92, precision of 0.92, recall of 0.92, accuracy of 0.97, and Matthews' Correlation Coefficient (MCC) of 0.90.

BERT also outperformed traditional models with respect to receiver operating characteristic (ROC) and precision-recall curves, achieving AUCs of 0.99 and 0.97 respectively (Figure \ref{fig:roc_lr}). BERT also fared better with differentiating between suicidal ideation and bereavement; BERT classified 30\% of bereavement posts as suicidal, whereas logistic regression classified 39\% as suicidal (See Appendix for subreddit-specific results). 


\section{Discussion}

\begin{figure}[h]
    \centering
    \includegraphics[width=0.42\textwidth]{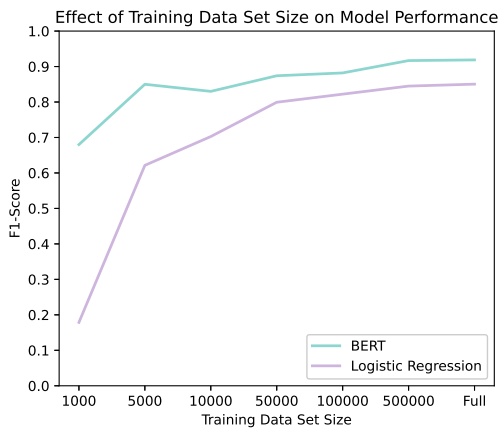}
    \caption{Both BERT and logistic regression achieve better performance when trained on more data.}
    \label{fig:size_importance}
\end{figure}

When trained on the Robin dataset, both traditional machine learning and large scale pre-trained language models offer state-of-the-art performance on the suicidal ideation detection task. It's worth noting that an instrumental component of achieving this performance was the size of the Robin dataset. As demonstrated in Figure \ref{fig:size_importance}, models trained on more data offered reliably better performance for both approaches; this demonstrates the importance and contribution of having such a large dataset. In the context of suicidal ideation, marginal increases in performance can result many additional lives each year saved through proactive intervention.

In addition to providing language models with a far more nuanced understanding of suicidal text, our dataset presents several other opportunities for further exploration. Robin has the potential to further psychological research into suicidal text online, an area of psychiatry which has thus far received disproportionately less attention than real-world incidences. The dataset could also be used to help train human annotators and clinicians who may have limited exposure to online suicidal ideation. We hope that Robin is used by the NLP community to power future online suicide prevention.


\section{Conclusion}

We presented the Robin dataset, a novel suicidal text corpus with noteworthy breadth and scale. Sourced from Reddit, Robin includes over 1.1 million posts and numerous categories of suicidal text, ranging from suicide bereavement to suicide awareness. These categories are intended to facilitate a better understanding of suicidal text by trained models. We demonstrated that Robin can be used to develop state-of-the-art suicidal sentiment classification models via both traditional machine learning and large scale pre-trained language models. This work marks a substantial step forward in the development of massive publicly available suicidal text datasets, as well as for highly accurate suicidal sentiment classification models.


\section{Ethical Considerations}

This paper requires ethical considerations on two fronts: the dataset itself, as well as the development and deployment of suicidal sentiment classification models. As discussed in the body of the paper, numerous efforts were made to maintain high ethical standards in the development of the Robin dataset. The corpus is anonymized and all Reddit post IDs were replaced with dataset-specific identifiers. It's also worth noting that the demographics of Reddit do not necessarily match those of the population at large. As a result, careful attention must be paid with respect to classification models, as performance may vary widely across different demographics. Additionally, mental health is oftentimes a sensitive subject, and it is of the utmost importance that any deployed models respect individual privacy and confidentiality. Based on our calculations and MTurk data, it took an average of approximately 10 seconds to annotate a post. Based on our payment per post, this converted to an hourly rate above U.S. federal and local state minimum wages.

To work with BERT, we rented a single Nvidia V100 GPU located in Northern Europe for a total of 35 hours. Based on efficiency data and quantification metrics presented in \citet{lacoste2019quantifying}, the estimated carbon output of this research amounted to approximately 0.53 kg $CO_2$ equivalent. It is important for researchers to be conscious of the environmental impact of their research.


\bibliography{custom}


\cleardoublepage
\appendix
\setcounter{table}{0}
\setcounter{figure}{0}
\renewcommand{\thefigure}{A\arabic{figure}}
\renewcommand{\thetable}{A\arabic{table}}

\begin{figure*}[h]
    \centering
    \includegraphics[width=1.0\textwidth]{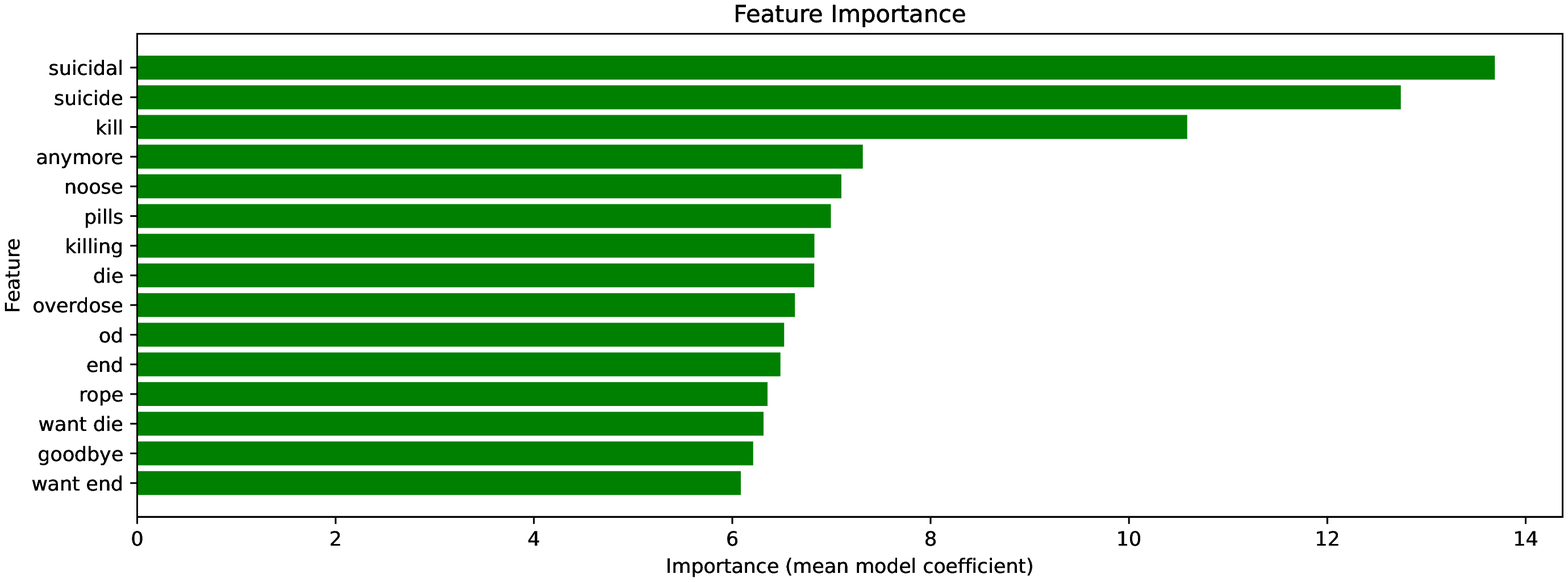}
    \caption{Salient features extracted from the Robin dataset by the logistic regression model via mean model coefficient.}
    \label{fig:importance_lr}
\end{figure*}

\begin{table*}[hbt!]
\centering
\begin{tabular}{lll}
\hline
\textbf{Subreddit} & \textbf{BERT} & \textbf{Logistic Regression}\\
\hline
    books & 0.23 & 0.21\\
    casualconversation & 1.06 & 1.57\\
    jokes & 0.31 & 1.01\\
    mensrights & 0.76 & 0.7\\
    politicaldiscussion & 0.04 & 0.19\\
    relationships & 0.47 & 0.8\\
    self & 7.32 & 8.02\\
    showerthoughts & 0.37 & 2.11\\
    suicidebereavement & 30.11 & 38.73\\
    suicidewatch & 91.83 & 81.56\\
    teenagers & 2.55 & 3.08\\
    television & 0.08 & 0.27\\
    tifu & 0.18 & 0.52\\
    truegaming & 0.08 & 0.19\\
    
\hline
\end{tabular}
\caption{\label{table:model_positives}
Percentage of each subreddit's posts classified as suicidal by BERT and logistic regression models trained on Robin.}
\end{table*}

\section{Extended Benchmarking Methods and Results}
\label{sec:subreddit_specific}
As specified in the body of the paper, we trained a variety of traditional machine learning models to benchmark results on the classification task for this dataset. All models were hyperparameter tuned via a grid search that also included six data pipelines: we used three tokenizers (unigrams, bigrams, unigrams and bigrams) with two vectorizers (bag of words, TF-IDF). Each model was trained in sci-kit learn with 5 cross validation folds using the entire dataset. 

Our LinearSVC models were trained with regularization parameters (C) of [$0.1$, 1, 10, 100], a stopping tolerence of $1\cdot 10^{-3}$, and with attempts for both the dual and primal optimizaiton problem. We achieved our best results, an F1-Score of 0.85, when we used LSVC with $C=0.1$, primal optimization, and unigrams and bigrams tokenization with TF-IDF vectorization.

Both our complement and multinomial Naive Bayes models had no tuning. We achieved an F1-Score of 0.811 with multinomial naive Bayes when combined with unigrams and bigrams using bag of words vectorization. We achieved an F1-Score of 0.805 with complement naive Bayes when combined with unigrams and bigrams using bag of words vectorization.

Our logistic regression model also had no tuning, using scikit-learn default parameters. We achieved an F1-Score of 0.85 when the model was combined with unigrams and bigrams using bag of words vectorization.

Our random forest models were tuned with max depth parameters of [5, 6, 8, 10] and numbers of estimators in the range of [50, 100, 250]. Our best results were with a max depth of 8 and 250 estimators, with which we achieved an F1-score of 0.729 when we combined this model with unigrams and bag of words vectorization.

Finally, we trained XGBoost with learning rates of [0.10, 0.30, 0.50], max depths of [3, 5, 7], minimum child weights of  [1, 3, 5], a gamma of 0.0, a colsample\_bytree of 1.0, and numbers of estimators in the range [50, 100, 200]. We achieved our best results with a learning rate of 0.30, a max depth of 7, a minimum child weight of 1, a gamma of 0.0, a colsample\_bytree of 1.0, and 200 estimators. With these parameters, we were able to achieve an F1-Score of 0.814 when we combined this model with unigrams and TF-IDF vectorization.

Figure \ref{fig:importance_lr} demonstrates the feature importance of our logistic regression model as determined by parameter coefficients. Additionally, Table \ref{table:model_positives} demonstrates the percentage of each subreddit that was classified as suicidal by the logistic regression model, as well as the finetuned BERT model.

\setcounter{table}{0}
\setcounter{figure}{0}
\renewcommand{\thefigure}{C\arabic{figure}}
\renewcommand{\thetable}{C\arabic{table}}

\end{document}